\documentclass[11pt]{article}
\usepackage{eamt24}
\usepackage{times}
\usepackage{url}
\usepackage{latexsym}
\usepackage[small,bf]{caption} 
\usepackage{cleveref}
\usepackage{tipa}
\title{ExU: AI Models for Examining Multilingual Disinformation Narratives and Understanding their Spread}

\author{
Jake Vasilakes\textsuperscript{1}, 
Zhixue Zhao\textsuperscript{1},
Ivan Vykopal\textsuperscript{2},
Michal Gregor\textsuperscript{2}, \\
\textbf{Martin Hyben}\textsuperscript{2},
\and
\textbf{Carolina Scarton\textsuperscript{1}} 
\\[0.3cm]
\textsuperscript{1} Department of Computer Science, University of Sheffield, UK \\
\textsuperscript{2}  Kempelen Institute of Intelligent Technologies, Bratislava, Slovakia\\ 
\texttt{\small \{j.vasilakes, zhixue.zhao, c.scarton\}@sheffield.ac.uk} \\
\texttt{\small \{michal.gregor, ivan.vykopal, martin.hyben\}@kinit.sk} 
}

\date{}

\begin{document}
\maketitle


\newcommand{\exufootnote}{\footnote{Exu ~/\textipa{ISu}/ is a deity associated with orderliness (Yoruba religion) or the guardian of houses and a messenger god (Afro-Brazilian religions). Exu has been wrongly associated with the devil in Western culture, with mis-translations of their name, often a victim of disinformation campaigns aiming to demonise African religions.}}

\section{Project Overview}

Online disinformation is a major challenge, with potential to cause economic, social, and medical harm \cite{Zubiaga_Aker_Bontcheva_Liakata_Procter_2018}.
Disinformation can be disseminated in multiple languages, which can be an overwhelming challenge
for fact-checkers and journalists. It is therefore necessary to develop multilingual methods for analysing 
disinformation. The ExU project\footnote{\url{https://exuproject.sites.sheffield.ac.uk}} aims to do just that, targeting stance classification and claim retrieval, two central tasks for assisting fact-checkers. 

Stance classification predicts whether a piece of content (e.g., a social media post or news article) agrees or disagrees with a claim. Claim retrieval aims to find relevant fact-checks for a given claim. Previous research in these areas, predominantly in English, is largely focused on single languages \cite{Kucuk_Can_2020}. Still, there is no research that focuses on developing and evaluating at large-scale a single stance detection or claim retrieval model for multiple languages.

Given these challenges, the objectives of the ExU project are to (1) develop novel methods for multilingual disinformation analysis via the tasks of stance detection and claim retrieval and (2) follow a multilingual user-centric evaluation which focuses on providing explainability of model predictions to end users.
Besides English, ExU will work with a set of 20+ languages, providing evaluation frameworks for Portuguese, Spanish, Polish, Slovak, Czech, Hindi and French (languages spoken in the UK and Slovakia). ExU started in November 2023 and is an 18-month project.

\section{Progress}

We conducted a survey of user requirements for our proposed tools at the Voices Festival of Journalism and Media Literacy,\footnote{\url{https://voicesfestival.eu/}} which brings together journalists, fact-checkers, researchers, and educators. 
Participants were recruited among the visitors to the EMIF (European Media and Information Fund) booth. 
The survey consisted of 24 questions covering basic demographic information, exposure to multiple languages, and features of stance classification and claim retrieval.

\begin{figure}[!ht]
    \centering
    \includegraphics[width=0.7\linewidth]{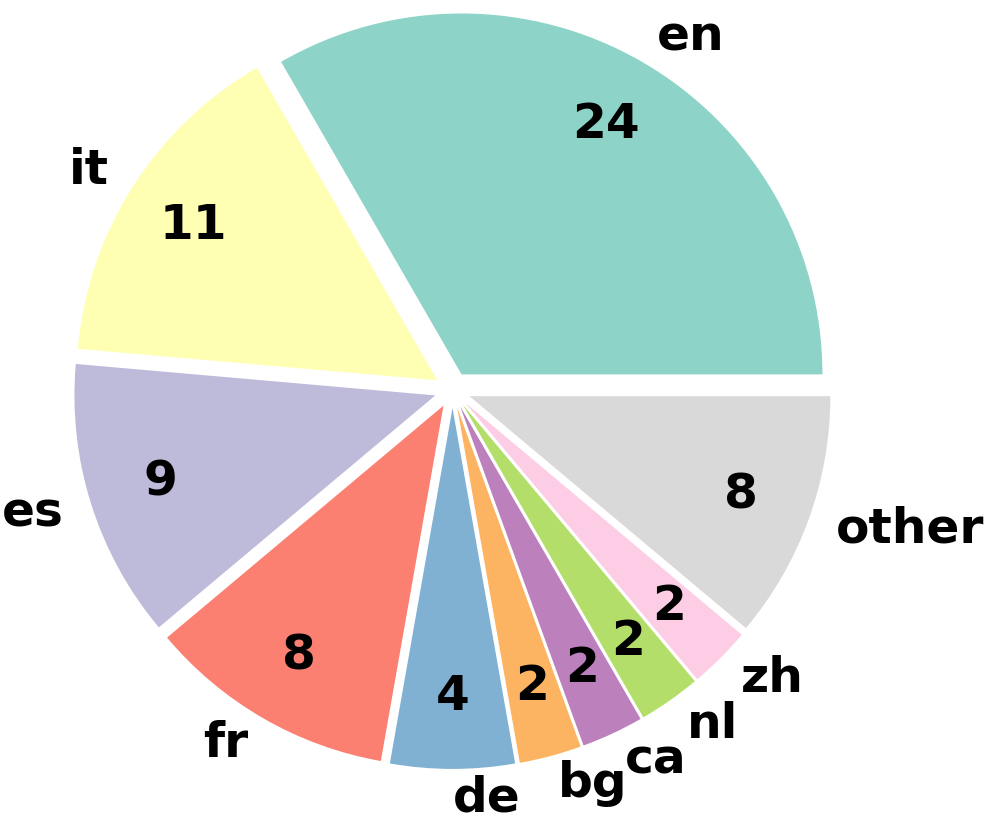}
    \caption{Counts of languages from responses to the survey question ``Which languages do you encounter most often in your work?''. The ``Other'' category is comprised of Czech, Hindi, Polish, Portuguese, Russian, Sinhala, Slovak, and Turkish, all of which had a count of one.}
    \label{fig:lang_counts}
\end{figure}

We obtained 29 survey responses. Almost all of participants (97\%) encountered content in multiple languages when performing fact-checks. Figure \ref{fig:lang_counts} indicates the counts of the languages from the participant's answers.\footnote{The event was held in Florence, Italy, so the results are biased towards EU languages and Italian specifically. We plan to obtain survey data from other demographics in the future.} For a fact-checking tool in general, participants would like content translated into a language of their choice, so it will be necessary to ensure accuracy of translations both between our target languages and into other user-specified languages. For stance detection, respondents deemed it most important to automatically predict the stance of the post regarding the target claim and to automatically highlight the main argument of posts. For claim retrieval, respondents would like a high-level summary of the claim's fact-checks in addition to the fact-checks themselves.

\vspace*{-2em}
\section{Future work}

\vspace*{-1em}
Based on the initial survey results, we aim for our stance classification models to output accurate and explainable predictions for content across the target languages. Towards this we will utilise multilingual transformers such as Aya \cite{ustun2024aya}, which covers all the languages in Figure \ref{fig:lang_counts}. To address the lack of data for low-resource languages we may obtain small amounts of target language fine-tuning data, as previous work found this improved results \cite{scarton2021cross}. Additionally, we may translate low-resource languages into English before performing classification to leverage the knowledge from English models. We are exploring explainability via feature attribution and rationale extraction, and our preliminary research shows promise for using extractive rationales in multiple languages. Still, explanations ought to be consistent across languages and invariant to translation, yet previous work showed a performance gap in explainability methods between mono- and multi-lingual models \cite{zhao-aletras-2023-incorporating}, so we plan to explore this in depth.


For multilingual claim retrieval, we aim to employ a retrieval augmented generation model \cite{Lewis_et_al_RAG_2020} to help end users efficiently discern factual claims from debunked ones. This model may facilitate the existing tools for extraction of textual claims from any textual content found online and match them with existing fact-checks contained in our MultiClaim dataset \cite{PikuliakSMHSMVS23}. The dataset contains 293,169 fact-checked articles and their corresponding claims in 39 languages. The output of the model will include the list of fact-checked claims relevant for each textual claim from the analysed textual content, their language and source references, along with the central claim summarisation of the retrieved claims in natural language.

\section{Acknowledgements}

The ExU project is funded by the European Media and Information Fund (grant number 291191). The sole responsibility for any content supported by the European Media and Information Fund lies with the author(s) and it may not necessarily reflect the positions of the EMIF and the Fund Partners, the Calouste Gulbenkian Foundation and the European University Institute.

\vspace*{-1.5em}
\bibliography{eamt24}

\begin{thebibliography}{}

\bibitem[\protect\citename{Küçük and Can}2020]{Kucuk_Can_2020}
Küçük, Dilek and Fazli Can.
\newblock 2020.
\newblock Stance detection: A survey.
\newblock {\em ACM Computing Surveys}, 53(1):12:1--12:37, February.

\bibitem[\protect\citename{Lewis \bgroup et al.\egroup }2020]{Lewis_et_al_RAG_2020}
Lewis, Patrick, Ethan Perez, Aleksandra Piktus, Fabio Petroni, Vladimir Karpukhin, Naman Goyal, Heinrich Küttler, Mike Lewis, Wen-tau Yih, Tim Rocktäschel, Sebastian Riedel, and Douwe Kiela.
\newblock 2020.
\newblock Retrieval-augmented generation for knowledge-intensive {NLP} tasks.
\newblock In {\em Advances in Neural Information Processing Systems}, volume~33, page 9459–9474. Curran Associates, Inc.

\bibitem[\protect\citename{Pikuliak \bgroup et al.\egroup }2023]{PikuliakSMHSMVS23}
Pikuliak, Mat{\'{u}}s, Ivan Srba, R{\'{o}}bert M{\'{o}}ro, Timo Hromadka, Timotej Smolen, Martin Melisek, Ivan Vykopal, Jakub Simko, Juraj Podrouzek, and M{\'{a}}ria Bielikov{\'{a}}.
\newblock 2023.
\newblock Multilingual previously fact-checked claim retrieval.
\newblock In Bouamor, Houda, Juan Pino, and Kalika Bali, editors, {\em Proceedings of the 2023 Conference on Empirical Methods in Natural Language Processing, {EMNLP} 2023, Singapore, December 6-10, 2023}, pages 16477--16500. Association for Computational Linguistics.

\bibitem[\protect\citename{Scarton and Li}2021]{scarton2021cross}
Scarton, Carolina and Yue Li.
\newblock 2021.
\newblock Cross-lingual rumour stance classification: a first study with {BERT} and machine translation.
\newblock In {\em Truth and Trust Online}, pages 50--59.

\bibitem[\protect\citename{Zhao and Aletras}2023]{zhao-aletras-2023-incorporating}
Zhao, Zhixue and Nikolaos Aletras.
\newblock 2023.
\newblock Incorporating attribution importance for improving faithfulness metrics.
\newblock In Rogers, Anna, Jordan Boyd-Graber, and Naoaki Okazaki, editors, {\em Proceedings of the 61st Annual Meeting of the Association for Computational Linguistics (Volume 1: Long Papers)}, pages 4732--4745, Toronto, Canada, July. Association for Computational Linguistics.

\bibitem[\protect\citename{Zubiaga \bgroup et al.\egroup }2018]{Zubiaga_Aker_Bontcheva_Liakata_Procter_2018}
Zubiaga, Arkaitz, Ahmet Aker, Kalina Bontcheva, Maria Liakata, and Rob Procter.
\newblock 2018.
\newblock Detection and resolution of rumours in social media: A survey.
\newblock {\em ACM Computing Surveys}, 51(2):32:1--32:36, February.

\bibitem[\protect\citename{Üstün \bgroup et al.\egroup }2024]{ustun2024aya}
Üstün, Ahmet, Viraat Aryabumi, Zheng-Xin Yong, Wei-Yin Ko, Daniel D'souza, Gbemileke Onilude, Neel Bhandari, Shivalika Singh, Hui-Lee Ooi, Amr Kayid, Freddie Vargus, Phil Blunsom, Shayne Longpre, Niklas Muennighoff, Marzieh Fadaee, Julia Kreutzer, and Sara Hooker.
\newblock 2024.
\newblock Aya model: An instruction finetuned open-access multilingual language model.
\newblock {\em arXiv preprint arXiv:2402.07827}.

\end{thebibliography}
\bibliographystyle{eamt24}
\end{document}